\documentclass[letterpaper, 10 pt, conference]{ieeeconf}  

\IEEEoverridecommandlockouts                              

\usepackage{graphicx} 
\usepackage{amsmath} 
\usepackage{amsfonts} 
\usepackage{dsfont}
\usepackage[font=small,labelfont=bf,tableposition=top]{caption}
\usepackage{mwe}
\usepackage{capt-of,etoolbox}
\usepackage{color}

\title{\LARGE \bf
Learning Terrain-Adaptive Locomotion with Agile Behaviors by Imitating Animals
}
\author{Tingguang Li, Yizheng Zhang, Chong Zhang, Qingxu Zhu, Jiapeng sheng, Wanchao Chi, Cheng Zhou, Lei Han
\thanks{$^{1}$Tingguang Li, Yizheng Zhang, Chong Zhang, Qingxu Zhu, Jiapeng sheng, Wanchao Chi, Cheng Zhou and Lei Han are with Tencent Robotics X Laboratory, China.
        {\tt\small \{teaganli, yizhenzhang, chongzzhang, qingxuzhu, kevinsheng, wanchaochi, mikechzhou, lxhan\}@tencent.com}.}%
}

\begin{document}
\maketitle
\thispagestyle{empty}
\pagestyle{empty}
\begin{abstract}
In this paper, we present a general learning framework for controlling a quadruped robot that can mimic the behavior of real animals and traverse challenging terrains. Our method consists of two steps: an imitation learning step to learn from motions of real animals, and a terrain adaptation step to enable generalization to unseen terrains. We capture motions from a Labrador on various terrains to facilitate terrain adaptive locomotion. Our experiments demonstrate that our policy can traverse various terrains and produce a natural-looking behavior. We deployed our method on the real quadruped robot \emph{Max}~\cite{Chi2022} via zero-shot simulation-to-reality transfer, achieving a speed of $1.1$ m/s on stairs climbing.
\end{abstract}

\section{INTRODUCTION}
\begin{figure*}[t]
    \centering
    \includegraphics[width=\textwidth]{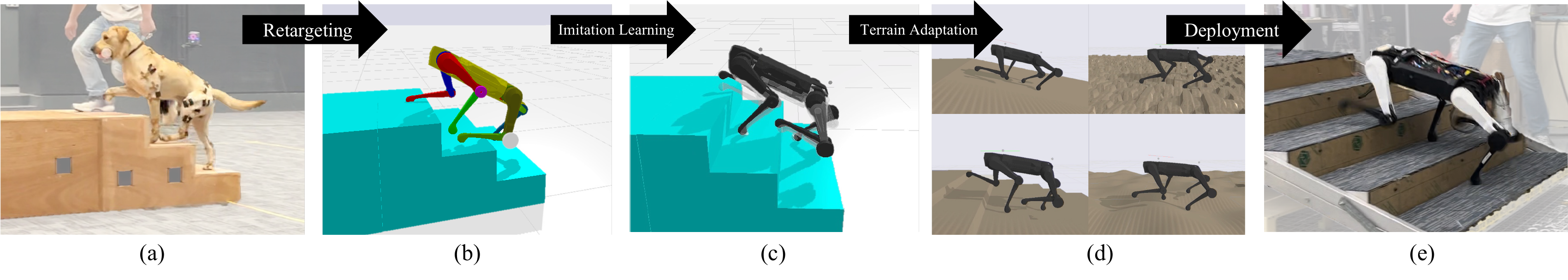}
    \caption{The pipeline of the proposed method. (a) We capture real dog motions with terrain data. (b) The motion data is retargeted to our robot skeleton. This figure shows the kinematic robot. (c) At the imitation learning step, the dynamic robot (the black one) tracks the reference motion (the transparent one). (d) The behavior generalizes to more unseen terrains. (e) We deploy our policy on a real robot.}
    \label{fig:pipeline}    
    \vspace{-0.3cm}
\end{figure*}

Quadruped locomotion over varied terrains is a challenging yet fascinating area of study. The four-legged support structure provides a great level of stability and maneuverability, making the quadruped robot capable of traversing complex, uneven, and unstructured environments.  
Early research in this field focused on developing controllers by approximating system dynamics and formulating an optimization problem to solve for the optimal actions for a given objective~\cite{Chi2022,Semini2011,Bellicoso2017,DiCarlo2018,Jiang2023}. However, these controllers typically require a specifically designed planner to achieve locomotion over rough terrains, significantly hindering their adaptability to 
unseen environments. 

Recently, Reinforcement Learning (RL) has achieved great successes in controlling quadruped robots~\cite{hwangbo2019learning,9982132,9982038}. RL requires only no or limited prior knowledge about the system and trains its policy in simulation before deploying it to the real world. However, due to the Simulation-to-Reality Gap (Sim-to-Real Gap), the policy often diverges in the real world from that in the simulation. To address this gap, one popular approach is privileged learning~\cite{doi:10.1126/scirobotics.abc5986,doi:10.1126/scirobotics.abk2822,margolisyang2022rapid,lai2022sim}, where an \emph{oracle} teacher policy is trained with privileged information that is inaccessible in the real world, and a student policy is trained to infer this information from the observation and reproduce the behavior. 
However, the resulting locomotion strategies are often idiosyncratic and unnatural compared to the versatility and efficiency of animal locomotion. Another category has been proposed to imitate motions from animals~\cite{peng2018deepmimic,RoboImitationPeng20,escontrela2022adversarial,10.1145/3450626.3459670,peng2022ase}, which can generate physically plausible and natural-looking behavior. However, the terrain adaptability of these policies has not been addressed yet.

In this paper, we focus on developing the robot's capability to traverse various challenging terrains while exhibiting natural behavior. The pipeline of our method is illustrated in Fig.~\ref{fig:pipeline}. First, we collect motion data from a Labrador on terrains including slopes and stairs. We then re-target this motion data to our quadruped robot skeleton. Our method comprises a two-step training process. In the imitation learning step, we train the robot to track motion clips via reinforcement learning in simulation. To enable the motion to generalize to more terrain types, we introduce a terrain adaptation step. We comprehensively evaluate our method in both simulation and reality. Our policy reaches a speed of $1.1$ m/s when climbing stairs with the \emph{Max} robot~\cite{Chi2022} in the real world. In conclusion, the primary contributions of this paper are:
\begin{itemize}
    \item We assemble a motion capture dataset from a real dog with motions on different terrains. To the best of our knowledge, our dataset is the first to contain terrain-aware motion data\footnote{We will release this dataset in the future.};
    \item We propose a general two-step learning process, producing a policy capable of traversing on various terrains and exhibiting natural behavior;
    \item We deploy our controller on a real quadruped robot, achieving successful zero-shot sim-to-real transfer and demonstrating it on diverse terrains.
\end{itemize}


\section{RELATED WORK}

The conventional control approaches for legged robots has long been an area of interests in robotics. Many of these approaches rely heavily on extensive manual engineering~\cite{raibert1984hopping,raibert1986legged,bledt2018cheetah}. In order to enable legged robots to traverse uneven terrains, complex state machines are designed to coordinate the execution of motion primitives and reflex controller~\cite{bledt2018contact,focchi2020heuristic}.

Model-free reinforcement learning has become an alternative approach for controlling robots, utilizing motion captured data from real animals to enable the robot to learn a wide range of skills without having to design behavior-specific reward functions~\cite{peng2018deepmimic,RoboImitationPeng20,10.1145/3450626.3459670,escontrela2022adversarial,vollenweider2022advanced,9981648,bohez2022imitate,2022-opt-mimic}. Peng et al.~\cite{peng2018deepmimic} adapted RL to learn robust control policies capable of imitating a variety of motion clips. During training, they combined a motion imitation objective with a task objective, allowing the robot to behave naturally and interactively. They applied their model to a quadruped robot~\cite{RoboImitationPeng20}, resulting in agile behaviors and natural-looking motions. Peng et al.~\cite{10.1145/3450626.3459670} further proposed the Adversarial Motion Prior (AMP) algorithm, which uses Generative Adversarial Imitation Learning (GAIL) to learn a style reward, obviating the need for manually designed objective functions. GAIL~\cite{ho2016generative} measures the similarity between the policy and the demonstration and optimizes the objective via RL. The objective is represented as a discriminator to predict whether a state is from the demonstration or the policy, and the policy is trained in an adversarial manner, making it difficult to distinguish between the policy's produced states and actions and the demonstration's. Escontrela et al.~\cite{escontrela2022adversarial} implemented AMP on a real quadrupedal robot, demonstrating that the style learned from a German Shepherd was capable of yielding energy-efficient locomotion strategies with natural gait transitions. Eric et al.~\cite{vollenweider2022advanced} proposed Multiple AMP, allowing for multiple discretely switchable styles. They tested their method with a wheeled-legged quadruped robot and showed skills such as switching between a quadrupedal and humanoid configuration. These works consider the natural behaviors, while the adaptation ability on terrains is less addressed. 

Previous works have employed various terrain adaptation techniques to deploy their robots on uneven terrains. Takahiro et al.~\cite{doi:10.1126/scirobotics.abc5986} proposed an automatic curriculum learning method to facilitate terrains adaptation. This method enables ANYmal robot to traverse over a variety of rough terrains. Kumar et al.~\cite{kumar2021rma} proposed a Rapid Motor Adaptation (RMA) algorithm to address the issue of real-time online adaptation in legged robots. The algorithm was trained on various terrain generators and deployed in difficult environments, including rocky, slippery, deformable surfaces. Won et al.~\cite{physicstogwon22} proposed a helper branch to facilitate agents on uneven terrains in the computer graphics field. Our work is inspired by this, and we introduce a terrain adaptation module with terrain adaptive parameter that we evaluate in reality.

The challenges of applying RL in real robots have proven to be the gap between simulation and the real world. To bridge this gap, several techniques have been proposed, including constructing more accurate simulators~\cite{tan2018sim,xie2020learning}, domain randomization~\cite{tobin2017domain,andrychowicz2020learning,peng2018sim}, and privileged learning~\cite{doi:10.1126/scirobotics.abc5986,doi:10.1126/scirobotics.abk2822,margolisyang2022rapid,lai2022sim,agarwal2022legged,kumar2021rma}. Privileged learning involves a teacher policy encoding privileged information (e.g. friction coefficients) which is not accessible in the real world, and then training a student policy to replicate the teacher's behavior. Lee et al.~\cite{doi:10.1126/scirobotics.abc5986} used a Temporal Convolutional Network (TCN) to implicitly reason about contact and slippage events from proprioceptive measurements. Miki et al.~\cite{doi:10.1126/scirobotics.abk2822} integrated exteroceptive and proprioceptive perception for legged locomotion with an attention-based recurrent encoder. Margolis et al.~\cite{margolisyang2022rapid} proposed an adaptive curriculum on velocity commands when training the teacher policy. Experiments on the MIT Mini Cheetah achieved speeds of up to 3.9 m/s. Lai et al.~\cite{lai2022sim} combined privileged learning with transformer through a two-stage training process: in the offline pretraining stage, the teacher policy interacts with a simulator and collects trajectories for transformer training, and in the online correction stage, transformer interacts with the simulator while the teacher simultaneously gives actions as the target, similar to Dagger~\cite{ross2011reduction}. This model was deployed on a real A1 robot and can traverse multiple terrains. 


\section{METHODOLOGY}

An overview of our pipeline is shown in Fig.~\ref{fig:pipeline}. First, we capture motion data from a Labrador using a Motion Capture system \emph{Vicon} and retarget it to the quadrupled robot \emph{Max}.
Our method consists of two steps: an imitation learning step and a terrain adaptation step, as shown in Fig.~\ref{fig:architecture}.
In the imitation learning step, we imitate all different reference motion clips using a variational autoencoder architecture. The encoder learns to map a sequence of future reference motion frames onto a latent embedding representing the target future movements. The decoder learns to maps this embedding, together with proprioceptive and exteroceptive information, to joint actuator commands. 
Then in the terrain adaptation step, we fix the decoder to maintain motion naturalness and train the policy in various terrains in a curriculum manner. This produces a robust policy that is capable of handling different challenging terrains with natural-looking behavior. 

\subsection{Motion Capture Data}
In imitation learning, the motion quality of robots is directly related to the reference motion data. 
Although there are some publicly available Motion Capture (MoCap) data~\cite{zhang2018mode}, they only contain motions on flat ground without considering terrain information.
We argue that the MoCap data with terrain is important to learn physically plausible motions on different terrains, such as stairs (which will be discussed in Section~\ref{sec:exp_simulation}).
Consequently, we have collected a MoCap dataset containing motion on both the plane and different terrains, including a stair and a slope.
Specifically, our dataset consists of motions including walking, running, and fast running on the ground, going up the stairs, going down the stairs, climbing up the slope, and descending down the slope. The details of the dataset are presented in TABLE~\ref{tab:dataset}. 
Our stair has three steps with the step width of $0.32$ m and step height of $0.16$ m.
The slope has an inclination of $15^{\circ}$. 

\begin{table}[h]
\caption{The breakdown of our dog motion dataset with terrains including stair up/down (StrUp/StrDown) and slope up/down (SlpUp/SlpDown).}
\label{tab:dataset}
\centering
\begin{tabular}{|c|c|c|c|c|c|}
\hline
         & Plane        & StrUp      & StrDown    & SlpUp  & SlpDown\\
\hline
 Walk (sec)   & 1121.6    & 149.7         & 146.4         & 202.8     & 149.7        \\
 Run (sec)    & 646.5     & 97.7         & 84.8       & 129.2     & 109.5       \\
 Fast Run (sec)   & 0         & 30.1          & 36.75       & 39.9      & 33.6        \\
\hline
\end{tabular}
\end{table}

\subsection{Motion Retargeting}
The MoCap data collected from a Labrador is morphologically different from our quadruped robot \emph{Max}. The \emph{Max} is wider with a lower base. 
To reduce this discrepancy, we retarget the dog MoCap data to the \emph{Max} robot using Inverse Kinematics (IK).
First, a set of keypoints (shoulder, shoulder blade, and haunch) from the dog is selected and used to compute the position and orientation of the base joint of the robot.w
Then, the foretoes and hindtoes from the dog are paired with the corresponding end effectors from the robot.
After the pose of the base and end effectors are determined, the rest of the robot's joints can then be solved using IK following~\cite{RoboImitationPeng20}.
To further reduce the morphology discrepancy, we lower the robot's base height and expand the legs on both sides, resulting in a more natural-looking behaviour of the robot.

\begin{figure}[t]
\includegraphics[width=\linewidth]{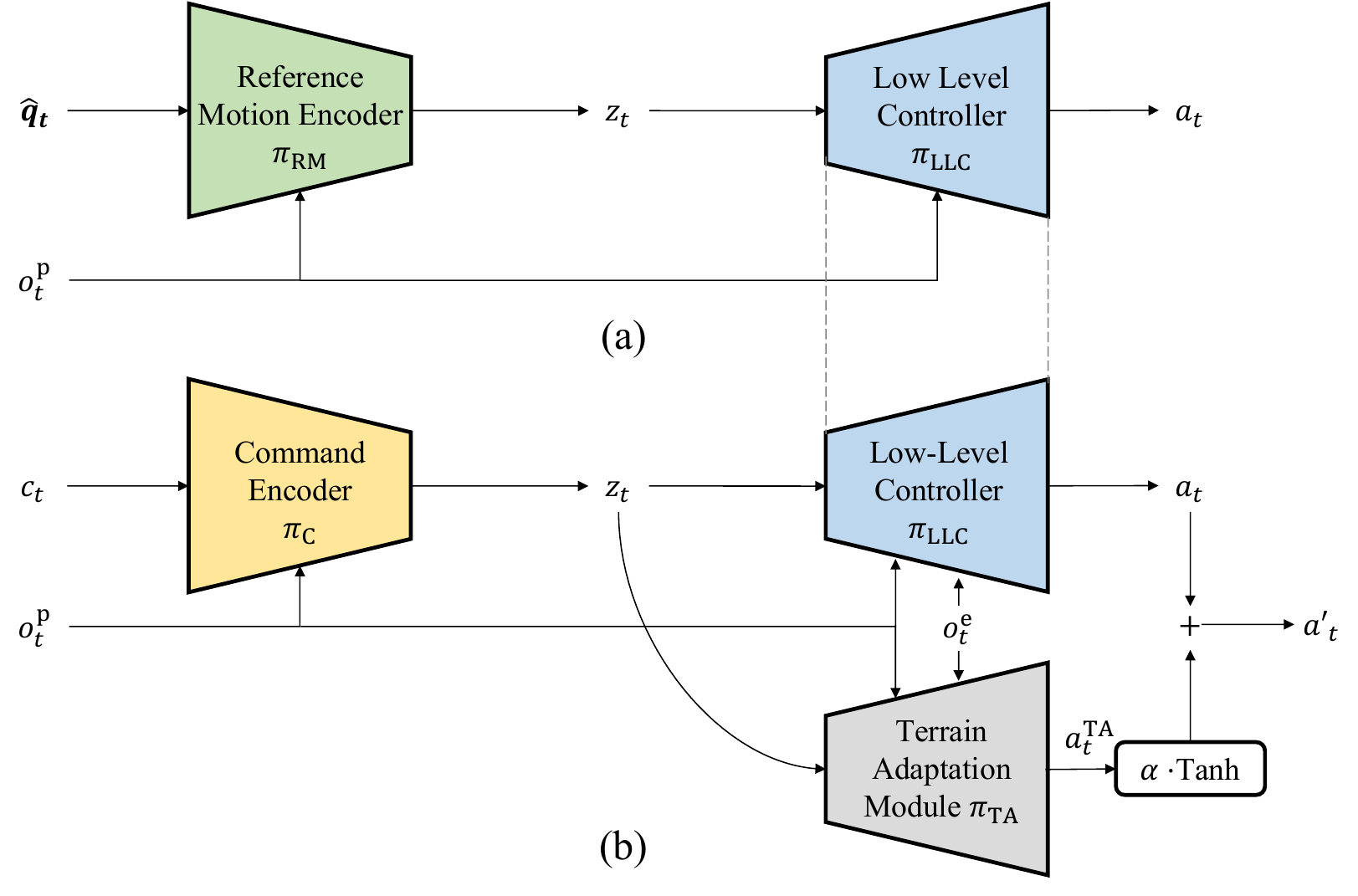}
\caption{The architecture overview. We have two training steps: (a) The imitation learning step: we train a reference motion encoder and a low-level controller to imitate motion clips; (b) The terrain adaptation step: we use an additional terrain adaptation module to learn to adapt to different terrains. The low-level controller is reused and kept fixed at this step (the blue part).}
\label{fig:architecture}
\vspace{-0.3cm}
\end{figure}

\subsection{Imitation Learning Step}
The goal of the imitation learning step is to train a single policy to imitate all the reference motion trajectories by optimizing
\begin{equation}
\arg\min_{a} \sum_{i,t}||\hat{s}_{i,t+1} - P(s_{t+1}|s_t, a_t)||_2^2,
\end{equation}
where $s$ represents the robot state including joint positions, orientations, velocities, etc. $\hat{s}_{i,t}$ is the state from the reference motion clip $i$ at frame $t$. $P$ represents the world dynamics model.
We formulate imitation learning as a reinforcement learning problem similar to~\cite{peng2018deepmimic}.
In reinforcement learning, the objective is to learn a policy $\pi$ to maximize its expected return. At each timestep, the agent receives a state $s_t$ and sample an action $a_t\sim \pi(a_t|s_t)$. The agent executes this action and receives a new state $s_{t+1}$ and reward $r_t$. In our task, the reward is designed to encourage the robot to track the reference motions.

Our policy is based on variational autoencoder architecture as shown in Fig.~\ref{fig:architecture}. 
The Reference Motion Encoder $\pi_{\text{RM}}(z_t|\hat{\textbf{\textit{q}}}_t,o^{\text{p}}_t)$ is conditioned on a sequence of future reference motions $\hat{\textbf{\textit{q}}}_t$ and proprioceptive information $o^{\text{p}}_t$, where $\hat{\textbf{\textit{q}}}_t = \{\hat{q}_{t+1/30}, \hat{q}_{t+1/15}, \hat{q}_{t+1/3}, \hat{q}_{t+1}\}$ represents target poses from the reference motion in the future $1/30$, $1/15$, $1/3$ and $1$ seconds respectively. $\pi_\text{RM}$ learns to map the future reference trajectories onto a latent embedding $z_t$.
We model the output of $\pi_{\text{RM}}(z_t|\hat{\textbf{\textit{q}}}_t,o^{\text{p}}_t)$ as a Gaussian distribution with mean $\mu^z_t$ and standard deviation $\sigma^z_t$ produced by out network
\begin{equation}
\pi_{\text{RM}}(z_t|\hat{\textbf{\textit{q}}}_t,o^{\text{p}}_t)=\mathcal{N}(\mu^z_t,\sigma^z_t).
\end{equation}
To ensure the latent space is well formed and a more focused exploration in the terrain adaptive step, we regularize the distribution of $\pi_\text{RM}$ by penalizing the KL divergence to a standard Gaussian, which is similar to conditional Variational Auto-Encoder (CVAE)~\cite{NIPS2015_8d55a249}:
\begin{equation}
\beta \mathbb{E}_{z\sim \pi_{\text{RM}}}[\text{KL}(\pi_{\text{RM}}(z_t|\hat{\textbf{\textit{q}}}_t,o^{\text{p}}_t)|p_\theta(z_t|\hat{\textbf{\textit{q}}}_t,o^{\text{p}}_t))],
\end{equation}
where $\beta=0.03$ and $p_\theta(z_t|\hat{\textbf{\textit{q}}}_t,o^{\text{p}}_t)$ is the prior distribution which is standard normal distribution in our case. The decoder, or the Low-Level Controller $\pi_{\text{LLC}}(a_t|z_t, o_t^\text{p}, o_t^\text{e})$ subsequently takes the latent variable $z_t$, proprioception observation $o_t^\text{p}$, and exteroception observation $o_t^\text{e}$, and produces the action $a_t$ for the robot. The low-level controller will be kept fixed and reused in the following terrain adaptation step.

The definition of reward function is similar to~\cite{peng2018deepmimic}, where the reward $r_t$ consists of $5$ terms: joint position reward $r^{\text{jpos}}_t$, joint velocity reward $r^{\text{jvel}}_t$, end effector position reward $r^{\text{epos}}_t$, base pose reward $r^{\text{bpose}}_t$, and base velocity reward $r^{\text{bvel}}_t$
\begin{equation}
r_t = 0.6\cdot r^{\text{jpos}}_t + 0.05\cdot r^{\text{jvel}}_t + 0.1\cdot r^{\text{epos}}_t + 0.15\cdot r^{\text{bpose}}_t + 0.1\cdot r^{\text{bvel}}_t,
\end{equation}
\begin{equation}
r^{\text{jpos}}_t = \text{exp}(-\sum\mathop{}_{\mkern-5mu j}||\hat{q}^j_t-q^j_t||^2),
\end{equation}
\begin{equation}
r^{\text{jvel}}_t = \text{exp}(-0.1\sum\mathop{}_{\mkern-5mu j}||\hat{\dot{q}}^j_t-\dot{q}^j_t||^2),
\end{equation}
\begin{equation}
r^{\text{epos}}_t = \text{exp}(-40\sum\mathop{}_{\mkern-5mu e}||\hat{p}^e_t-p^e_t||^2),
\end{equation}
\begin{equation}
r^{\text{bpose}}_t = \text{exp}(-20||\hat{p}^\text{base}_t-p^\text{base}_t||^2 - 10||\hat{q}^\text{base}_t-q^\text{base}_t||),
\end{equation}
\begin{equation}
r^{\text{bvel}}_t = \text{exp}(-2||\hat{\dot{p}}^\text{base}_t-\dot{p}^\text{base}_t||^2 -0.2||\hat{\dot{q}}^\text{base}_t-\dot{q}^\text{base}_t||^2),
\end{equation}
where $q^j_t$, $\dot{q}^j_t$ represent the 1D locol rotation and angular velocity of joint $j$ at time $t$ of the robot. $p^e_t$ denotes end-effector positions. $p^\text{base}_t$, $q^\text{base}_t$ are base position and base orientation, while $\dot{p}^\text{base}_t$ and $\dot{q}^\text{base}_t$ denote base linear velocity and angular velocity. $\hat{(\cdot)}$ represents the reference motion. 

\subsection{Terrain Adaptation Step}
The policy trained in the imitation learning step is capable of tracking reference motions, but there are some issues to be addressed before deploying it in the real world. 
Firstly, the imitation learning policy can only track reference motions with specific terrain configurations from our MoCap dataset, as the variance of terrains in the MoCap dataset is limited. Our dataset contains fixed size slopes and stairs, and thus tracking such reference motion in reality on terrains with different configurations may lead to failure. For example, tracking the motion of climbing stairs with $3$ steps would be very likely to hit the stair when there are more steps in real. Considering the fact that it would be very costly to collect large amounts of MoCap data with a wide variety of terrains, so we aim to make our policy generalize to various unseen terrains with the limited terrains present in our dataset.
Secondly, the policy is only able to track pre-defined trajectories without human control. We hope to make the policy controllable given target angle and speed.

To this end, we propose a terrain adaptation step. The objective of this step is to extend the policy to a variety of terrains, under given commands. For this step, the Reference Motion Encoder is replaced with Command Encoder $\pi_\text{C}(z_t|c_t,o^{\text{p}}_t)$, which encodes the user command $c_t$ and proprioception observation $o^{\text{p}}_t$. The user command $c_t=(\text{cos}(\hat{\theta}_t-\theta_t), \text{sin}(\hat{\theta}_t-\theta_t), \hat{v}_t)$, where $\hat{v}_t$, $\hat{\theta}_t$ and $\theta_t$ denote target speed, target yaw angle and robot yaw angle respectively. The reward function of this task is 
\begin{equation}
r_t = \text{exp}(-|\hat{v}_t-v_t|)\cdot \text{exp}(5\cdot (\text{cos}(\hat{\theta}_t - \theta_t)-1)),
\end{equation}
where $v_t$ is the 1D robot speed projected on the target robot direction.

To facilitate the policy performing on different terrains, we simultaneously train the agent using a curriculum learning approach on $7$ different terrains. The details of terrain curriculum learning are introduced in Section~\ref{sec:exp_setup}. Since the generated terrains do not necessarily match those from our dataset, this discrepancy may degrade the performance. As a solution, we can update the low-level controller simultaneously to adapt to unseen situations and improve the performance; however, this can lead to a \emph{forgetting} problem, where the motions become unnatural, and no longer resemble the real dog. To maintain the naturalness of the motion, we keep the low-level controller fixed during training, and propose the Terrain Adaptation module $\pi_\text{TA}$ (see Fig.~\ref{fig:architecture}) to allow the policy to adapt to new environments while retaining the original motion style. The command encoder $\pi_\text{TA}(a^\text{TA}_t|z_t,o^\text{p}_t,o^\text{e}_t)$ generates an action offset $a^\text{TA}_t$ which is added to the low-level controller, improving the performance in unseen terrains:
\begin{equation}
a_t' = a_t + \alpha \cdot \text{tanh}(a^\text{TA}_t),
\end{equation}
where $\alpha$ controls how much the terrain adaptive module affects the action. There is a trade-off between adaptability and naturalness: when $\alpha$ is small, the policy maintains its original motion style but performs poorly on challenging terrains, while when $\alpha$ is large, the robot performs well on terrains but loses its naturalness. We observe that the terrain adaptive module is essential for challenging terrains, while for flat surfaces, a low-level controller can handle it adequately. Therefore, we propose a terrain adaptive parameter $\alpha$ to control the extent to which the terrain adaptive module influences the action:
\begin{equation}
    \alpha = \begin{cases}
    0.1, &\text{if std}(o^{\text{e}}_t) > 0.01;  \\
    0, &\text{otherwise.}
\end{cases} 
\end{equation}
In this way, the robot can adapt to new terrains while maintaining it original motion style.


\section{EXPERIMENTS}\label{sec:exp}
\begin{table*}[t]
\caption{The comparison results of avarage return on different terrains in simulation. The results are averaged over $1000$ episodes with varying target angle, target speed and terrain parameters. We compare our method (Ours), Rhythmic Locomotion (RhyLoc)~\cite{sheng2022bio}, End-to-End RL (E2E RL), our method without Terrain Adaptation Module (w/o TAM) and our method without MoCap Terrain data (w/o MT). Bold numbers indicate the best scores.}
\label{tab:return}
\centering
\begin{tabular}{|c|ccccccc|}
\hline
                & Plane             & Slope Up          & Slope Down    & Stair Up      & Stair Down    & Block & Hill \\
\hline
Ours            & $323.88\pm 30.16$ & \textbf{347.51 $\pm$ 35.26} & \textbf{359.06 $\pm$ 13.85} & \textbf{355.21 $\pm$ 54.90} & \textbf{356.73 $\pm$ 16.22} & \textbf{352.76 $\pm$ 45.67} & \textbf{353.91 $\pm$ 47.41} \\
RhyLoc          & $125.56\pm 47.23$ & $218.13\pm 22.84$ & $191.51\pm 26.50$ & $217.38\pm 23.18$ & $224.50\pm 25.55$ & $216.63\pm 31.53$ & $209.47\pm 42.41$ \\
E2E RL   & \textbf{339.74 $\pm$ 22.59} & $302.29\pm 69.47$ & $354.58\pm 15.31$ & $342.31\pm 44.42$ & $354.21\pm 21.44$ & $322.39\pm 96.20$ & $333.48\pm 68.90$ \\
w/o TAM         & $305.94\pm 37.89$ & $296.12\pm 63.95$ & $343.65\pm 24.15$ & $262.75\pm 93.80$ & $329.20\pm 48.02$ & $235.80\pm 108.66$ & $273.36\pm 106.95$ \\
w/o MT  & $331.35\pm 28.15$ & $300.08\pm 52.87$ & $335.32\pm 41.65$ & $318.02\pm 66.19$ & $322.26\pm 60.84$ & $316.25\pm 85.87$ & $331.08\pm 24.01$ \\
\hline
\end{tabular}
\end{table*}
In this section, we report experimental results regarding the following questions:
(i) How effective is our terrain adaptation step and terrain adaptation module in improving the locomotion capability on challenging terrains such as slopes, stairs, blocks and hills?
(ii) How closely does our locomotion behavior resemble MoCap data from a real dog?
(iii) How robust is our approach when deployed in the real world? 
We compare our approach to an end-to-end baseline, and perform a comprehensive ablation study on different parts of our model.
Our hypothesis is that our approach is capable of learning motion styles of a real dog from MoCap data and exhibiting a stable locomotion on challenging unseen terrains.

\begin{table}[t]
\caption{Terrain curriculum learning parameters. Slope Inclination (SlpInc), Stair Step Height(StrHt), Step depth (StrDp), Block Size (BlkSz), Block Max Height (BlkHt), and Hill Height (HlHt).}
\label{tab:terrain}
\centering
\begin{tabular}{|c|cccccc|}
\hline
        & SlpInc    & StrHt     & StrDp     & BlkSz     & BlkHt     & HlHt  \\
\hline
Start   & $\pm$0.1  & 0         & 0.4       & 0.05      & 0.02      & 0.05  \\
End     & $\pm$0.4  &$\pm$0.15  & 0.34      & 0.15      & 0.1       & 0.2   \\
Step    & $\pm$0.02 &$\pm$0.01  & -0.005    & 0.03      & 0.01      & 0.02  \\
\hline
\end{tabular}
\end{table}

\subsection{Experimental Setup}\label{sec:exp_setup}
\textbf{Terrain Curriculum Learning}.
We implement our approach in PyBullet~\cite{coumans2016pybullet} to simulate the process of interacting with the environment, which include a variety of terrains.
In the imitation learning step, we reproduce the stairs and slope to match the scene used to collect the dog motion data. Therefore, thus allowing the MoCap data to fit the terrain well.
In the terrain adaptation step, we conduct terrain curriculum learning on $7$ terrains simultaneously, namely \emph{Plane}, \emph{Slope Up}, \emph{Slope Down}, \emph{Stair Up}, \emph{Stair Down}, \emph{Blocks} and \emph{Hills}. The slope inclination, stair step height, step depth, step number, block size, block maximum height and hill height are gradually increased according to a fixed curriculum, the details of which are presented in TABLE~\ref{tab:terrain}. Each episode begins at an easy setting and increases the difficulty after the robot succeeds for a few successive times. After reaching the maximum difficulty, the terrain is randomly sampled to prevent it from forgetting the easy ones.

\textbf{Observation space and action space}.
The observation includes proprioceptive information $o^\text{p}_t$ and exteroceptive information $o^\text{e}_t$. 
The proprioceptive observation $o^\text{p}_t$ consists of base linear and angular velocities, joint positions, joint velocities, last action and the gravity vector expressed in the IMU frame~\cite{barasuol2013reactive}. To capture the temporal information, we stack the previous $3$ proprioceptive observation such that $o^\text{p}_t \in \mathbb{R}^{135}$.
The exteroceptive observations $o^\text{e}_t\in \mathbb{R}^{1024}$ is a local height map centered at the robot. More specifically, $o^\text{e}_t$ includes $64\times 16$ pixels representing a $1.0\mbox{ m}\times 0.5\mbox{ m}$ rectangle patch. 
The latent variable $z_t \in \mathbb{R}^{8}$.
The action $a_t\in \mathbb{R}^{12}$ specifies the target positions for each joint, which is then converted to joint torques through a PD controller. Our policy generates action at the frequency of $50$Hz and the PD controller runs at $500$Hz.

\textbf{Domain Randomization}.
To facilitate robust sim-to-real transfer, we randomize terrain friction, actuator torque limit, and the masses of the robot's base in each episode. Additionally, we add noise to the exteroceptive observation and body center of mass estimation at each timestep. The domain randomization parameter distributions are summarized in TABLE~\ref{tab:domain}.
\begin{table}[t]
\caption{Domain Randomization Parameters}
\label{tab:domain}
\centering
\begin{tabular}{|c|c|}
\hline
Terrain friction & $U[0.5, 1.2]$ \\
Actuator torque limit & $U[16, 23]$ \\
Base mass multiplier & $U[0.7, 1.3]$ \\
Perception noise & $\mathcal{N}(0, 0.01)$ \\
Body Center of Mass & $U(-0.01, 0.01)$ \\
\hline
\end{tabular}
\vspace{-0.3cm}
\end{table}

\begin{figure}[t]
\includegraphics[width=\linewidth]{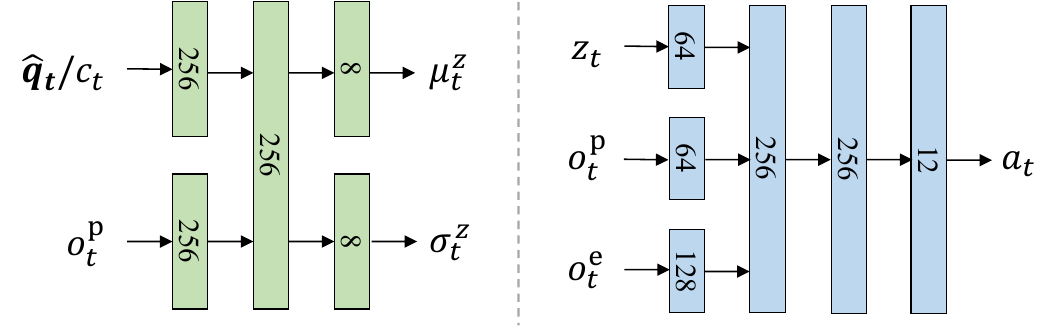}
\caption{The network details. The left green part shows the reference motion encoder $\pi_\text{RM}$ and command encoder $\pi_\text{C}$, with the input of $\hat{\textbf{\textit{q}}}_t$ and $c_t$ respectively. The right blue part shows the low level controller $\pi_\text{LLC}$ and terrain adaptation module $\pi_\text{TA}$. The numbers indicate the hidden unit dimensions.}
\vspace{-0.3cm}
\label{fig:network}
\end{figure}

\textbf{Model Representation}.
We parameterize our policy as MLP. 
The network structures are present in Fig.~\ref{fig:network}. The reference motion encoder $\pi_\text{RM}$ and command encoder $\pi_\text{C}$ share the same network structure (the green part), while the low level controller $\pi_\text{LLC}$ and terrain adaptation module $\pi_\text{TA}$ have the same network structure (the blue part).
For utilizing temporal information, we tried temporal models like LSTM, but it did not provide any additional reward and increased the training time in our task. Therefore, we opt for MLP with stacked adjacent frames as temporal information.

\subsection{Quantitative Evaluation in Simulation} \label{sec:exp_simulation}
We train the proposed approach using \emph{Proximal Policy Optimization} (PPO)~\cite{schulman2017proximal} under the distributed RL infrastructure \emph{TLeague}~\cite{sun2020tleague} with $1000$ parallel environments. 
An episode ends when the roll angle exceeds $45^\circ$ or is less than $-45^\circ$, the pitch angle is greater than $60^\circ$ or less than $-60^\circ$, or any robot parts other than toes make contact with the terrain. 
In the terrain adaptation step, to further discourage the robot from stepping on the edge of the stairs, we add an additional penalty when the toes are too close to the stair edge:
\begin{equation}
r^\text{stair}_t=-\sum_{i=1}^4 \mathds{1}_i,
\end{equation}
where
\begin{equation}
\mathds{1}_i= \begin{cases}
0.25, &\text{if Distance}(c_i, E) < 5\text{cm};  \\
0, &\text{otherwise.}
\end{cases} 
\end{equation}
where $c_i$ is the contact point of the i-th toe, and $E$ represents the closest stair edge to the i-th toe.

We evaluate the performance of our method on the task of following a given target speed and angle. 
The baseline model is the Rhythmic Locomotion (RhyLoc) method~\cite{sheng2022bio}, which incorporates a rhythm generator to stimulate periodic motor patterns. RhyLoc involves a sophisticated manually-designed reward function composed of $22$ terms, the majority of which aim to regularize the undesired behaviors. 
In comparison, our method only has 1 or 2 reward terms (for the stair environment), significantly simplifying reward shaping process.

To comprehensively evaluate the effectiveness of our method, we compare its performance with that of without imitation learning step (which is equivalent to the End-to-End RL, E2E RL), without the Terrain Adaptation Module (w/o TAM), and without MoCap Terrain data (w/o MT).
We run $1000$ episodes for each method, randomly sampling the target angle, target speed, and terrain parameters at each episode. TABLE~\ref{tab:return} presents the average return on different terrains. 
It is evident that our model achieves the highest scores on all the six challenging terrains, demonstrating the locomotion capability of our model.
The return of RhyLoc is lower than that of other methods, due to its reward terms which are designed to smooth out its actions. This makes a more stable policy and beneficial for real-world deployments, albeit at the cost of agility.
The E2E RL method optimizes its policy with the objective of maximizing the accumulated return, and thus achieves the highest score on the plane and the second highest score on the other six terrains. However, the resulting behavior is unnatural due to the lack of MoCap data guidance. For example, the robot learns to jump over different terrains instead of walking. Adding more regularization terms would help mitigate such effects, but this would also requires more human efforts to be involved.
The terrain adaptation module plays a significant role on the difficult terrains such as Stair Up, Slope Up, Block and Hill, by comparing our model with w/o TAM. This demonstrates its ability to adjust the motion to better fit challenging or unseen terrains.
For w/o MT, its performance is worse than ours, indicating the importance of the MoCap data with terrain for producing physically plausible motions. 
Further details on the motion behaviors can be found in our supplementary video.

To evaluate the naturalness of our method, we quantitatively compare the motion styles by evaluating four parameters:
\begin{itemize}
  \item Cycle time ($T_{\text{cycle}}$): length of the gait cycle in time;
  \item Swing Time ($T_{\text{swing}}$): time from toe off to heel strike;
  \item Stance Time ($T_{\text{stance}})$: time from heel strike to toe off;
  \item Step Distance ($D_{\text{step}}$): distance of the opposite side limbs heel strike in the direction of movement.
\end{itemize}
We evaluate our method (Ours), Rhythmic Locomotion (RhyLoc), and End-to-End Reinforcement Learning (E2E RL) on plane, slope, and stair surfaces. We also calculate these parameters from MoCap data. The results are reported in TABLE~\ref{tab:style}. It is clear that our model is quite close to the real dog data. 
For RhyLoc and E2E RL models, we observe that the cycle time and step distance are much shorter than the MoCap data, indicating that RL has learned a policy with denser cycles to better support the robot base. In comparison, our method learns the motion pattern from real dog data, resulting in a more natural behaviour. 

\begin{figure}
    \centering
    \includegraphics[width=200pt]{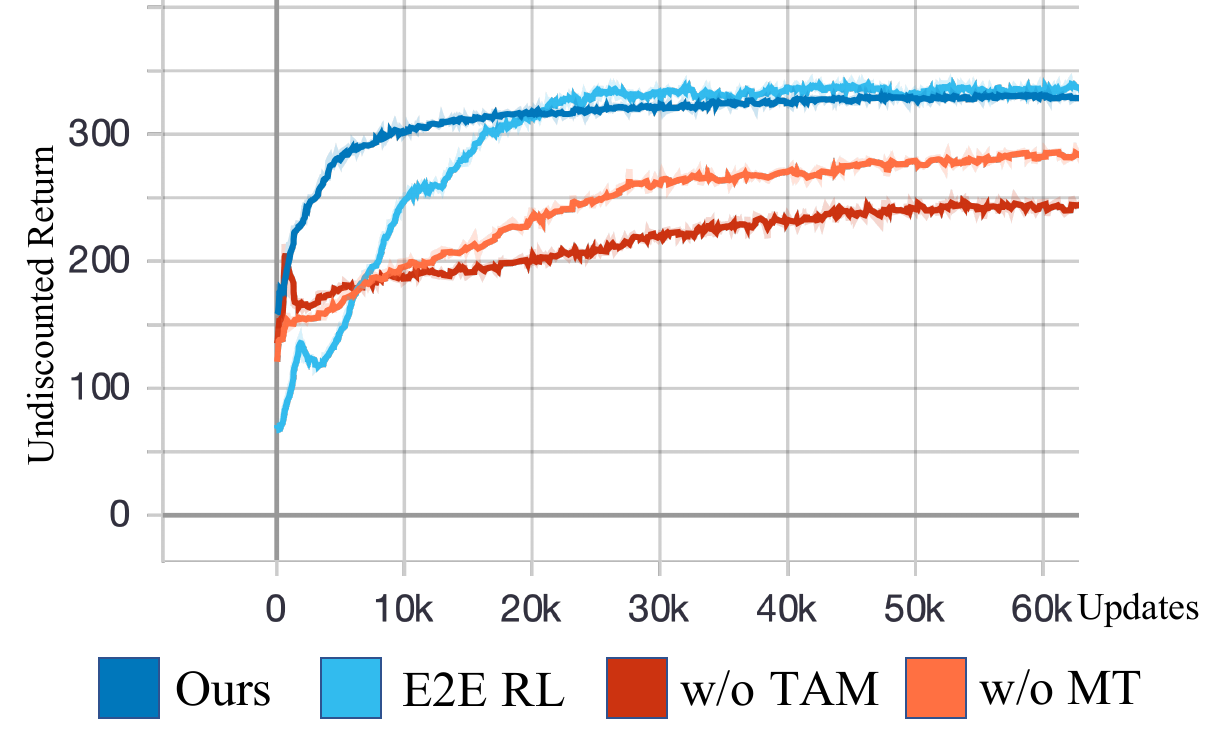}
    \caption{Learning curves for our approach, End-to-End RL (E2E RL), our method without Terrain Adaptation Module (w/o TAM) and our method without MoCap Terrain data (w/o MT).}     
    \label{fig:curves}
\end{figure}

We also report the training process as illustrated in Fig.~\ref{fig:curves}. It is evident that the curves of models without terrain adaptation module and MoCap terrain data learns more slowly than that of our proposed method. 
The terrain adaptation module generates a joint position residual that is added to the output of the low-level controller, thus allowing the motion to adapt to different terrains.
For MoCap terrain data, the low-level controller can learn to climb on stairs and slopes, which makes it easier to generalize across other terrains with different parameters. 

\begin{table}[h]
\caption{Style parameters of MoCap Data (Baseline), our approach (Ours), Rhythmic Locomotion (RhyLoc) and End-to-End RL (E2E RL). Bold numbers indicate the closest scores among Ours, RhyLoc and E2E RL to Mocap Data.}
\label{tab:style}
\centering
\begin{tabular}{|c|c|ccc|}
\hline
                &    MoCap Data & Ours             & RhyLoc  & E2E RL\\
\hline
$T_{\text{cycle}}$(s)   & $0.64\pm 0.16$ & \textbf{0.59 $\pm$ 0.07} & $0.43\pm 0.03$ & $0.30\pm 0.13$ \\
$T_{\text{swing}}$(s)   & $0.25\pm 0.04$ & \textbf{0.24 $\pm$ 0.05} & $0.17\pm 0.02$ & $0.15\pm 0.09$ \\
$T_{\text{stance}}$(s)  & $0.38\pm 0.11$ & \textbf{0.35 $\pm$ 0.06} & $0.26\pm 0.02$ & $0.16\pm 0.09$ \\
$D_{\text{step}}$(m)    & $0.31\pm 0.06$ & \textbf{0.29 $\pm$ 0.06} & $0.19\pm 0.06$ & $0.24\pm 0.14$ \\
\hline
\end{tabular}
\end{table}

\subsection{Qualitative Evaluation in Real Robot}
\begin{figure}
    \centering
    \includegraphics[width=250pt]{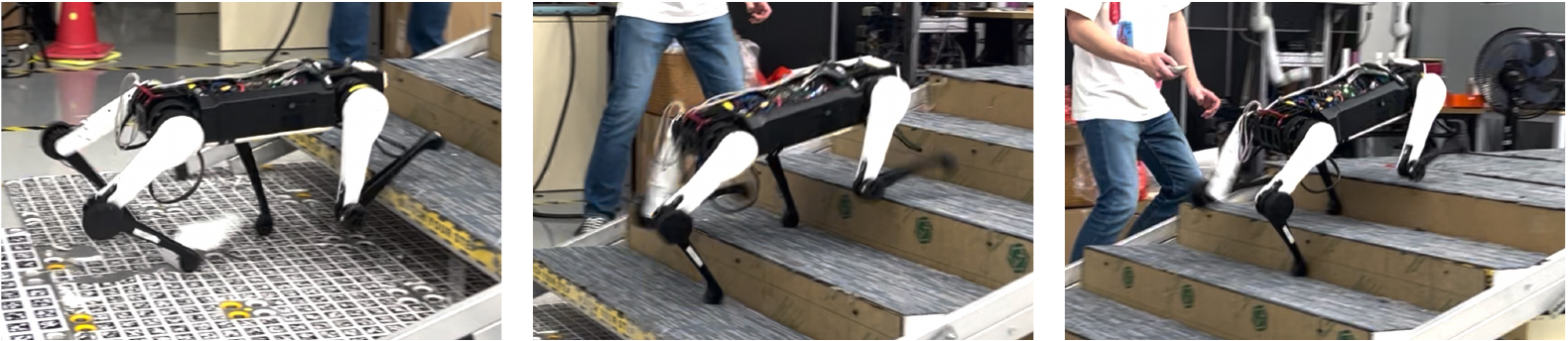}
    \caption{Snapshots of our method traversing stair up terrain. Our method achieves a speed of $1.1$m/s on the stair up terrain.}     
    \label{fig:deploy}
\end{figure}
We apply our method to the \emph{Max} robot in the real world without any fine-tuning. Although perception is essential for robots, we focus on the imitation and terrain adaptation part and thus use an \emph{Vicon} Motion Capture system to acquire the exteroceptive information directly. Specifically, we design a world height map that is the same as the terrain height in the real world, and then locate the robot inside the map using \emph{Vicon}. This enables us to collect terrain samples around the robot. 
All proprioceptive information are measured by sensors on the robot. We test our method on planes, slopes, and stairs. For the stair terrain, the first step height is $18$ cm and the rest $13$ cm. 
Our policy results in fast and agile behavior, comparable to animals, with a speed of $1.1$ m/s on the stair up terrain, as demonstrated in Fig.~\ref{fig:deploy}. To the best of our knowledge, this is the fastest speed achieved by a \emph{Max} robot climbing stairs.

\section{Conclusion}
In this paper, we present a two-step framework for terrain adaptive locomotion. Our method displays natural-looking behavior while being capable of traversing multiple terrains in simulation and in reality. However, our work has some limitations. Our assumption that exteroception terrain information is readily available is too strong, and our experiment was limited by the usage of a motion capture system. This prevents us from carrying out outdoor experiments. To address this, we could incorporate LiDAR or a camera into our framework. Additionally, out method does not address motion diversity, though it is able to perform multiple velocities with distinct behaviors. To further motion diversity distinction, we are considering involving more motion types such jumping and sitting.





\bibliographystyle{IEEEtran}
\bibliography{IEEEabrv}

\end{document}